\documentclass{article}
\usepackage{spconf,amsmath,graphicx}

\usepackage{multirow}
\usepackage{tabularx}
\usepackage{adjustbox}
\usepackage{subfig}
\usepackage{hyperref}

\usepackage{multirow}
\usepackage{pbox}
\usepackage{colortbl}
\usepackage{xcolor}
\usepackage{times}
\usepackage{epsfig}
\usepackage{graphicx}
\usepackage{amsmath}
\usepackage{amssymb}
\usepackage{makecell}
\newcommand{\comment}[1]{}

\usepackage{xcolor}
\usepackage{wrapfig}
\usepackage{tabularx}
\usepackage{subfig}
\usepackage{adjustbox}

\newcommand{\etal}{\textit{et al}.}

\newcommand{\done}[1]{}

\title{EGOK360: A 360 EGOCENTRIC KINETIC HUMAN ACTIVITY VIDEO DATASET}
\name{Keshav Bhandari$^1$, Mario A. DeLaGarza$^1$, Ziliang Zong$^1$, Hugo Latapie$^{2}$, Yan Yan$^1$}
\address{$^1$Department of Computer Science, Texas State University, USA\\
$^2$Chief Technology \& Architecture Office, Cisco, USA
}

\begin{document}
%
\maketitle
\begin{abstract}
Recently, there has been a growing interest in wearable sensors which provides new research perspectives for 360$^{\circ}$ video analysis. However, the lack of 360$^{\circ}$ datasets in literature hinders the research in this field. To bridge this gap, in this paper we propose a novel Egocentric (first-person) 360$^{\circ}$ Kinetic human activity video dataset (EgoK360). The EgoK360 dataset contains annotations of human activity with different sub-actions, \textit{e.g.}, activity Ping-Pong with four sub-actions which are pickup-ball, hit, bounce-ball and serve. To the best of our knowledge, EgoK360 is the first dataset in the domain of first-person activity recognition with a 360$^{\circ}$ environmental setup, which will facilitate the egocentric 360$^{\circ}$ video understanding. We provide experimental results and comprehensive analysis of variants of the two-stream network for 360 egocentric activity recognition. The EgoK360 dataset can be downloaded from \url{https://egok360.github.io/}.
\end{abstract}
\section{Introduction}
\label{sec:intro}
Wearable devices like Apple smartwatch, GoPro and Google Clip, have been widely used in our daily life nowadays. Meanwhile, the appearance of 360$^{\circ}$ cameras and the growing services on social media platforms such as Facebook and YouTube are changing the way how we consume multimedia. Having the advantage of 360 field-of-view over perspective videos from traditional cameras, 360$^{\circ}$ cameras have the superiority in many applications such as  self-driving cars, virtual-reality, life-logging, augmented reality, film-making and surveillance~\cite{sd1,sd2}. The popularity of 360$^{\circ}$ videos is also changing computer vision and virtual reality research area recently. Egocentric Activity Recognition (EAR) from videos is one of such fields. However, to the best of our knowledge, there is no public 360 egocentric human activity dataset in literature. 

\begin{figure}[htb]
\begin{center}
\includegraphics[width=\linewidth]{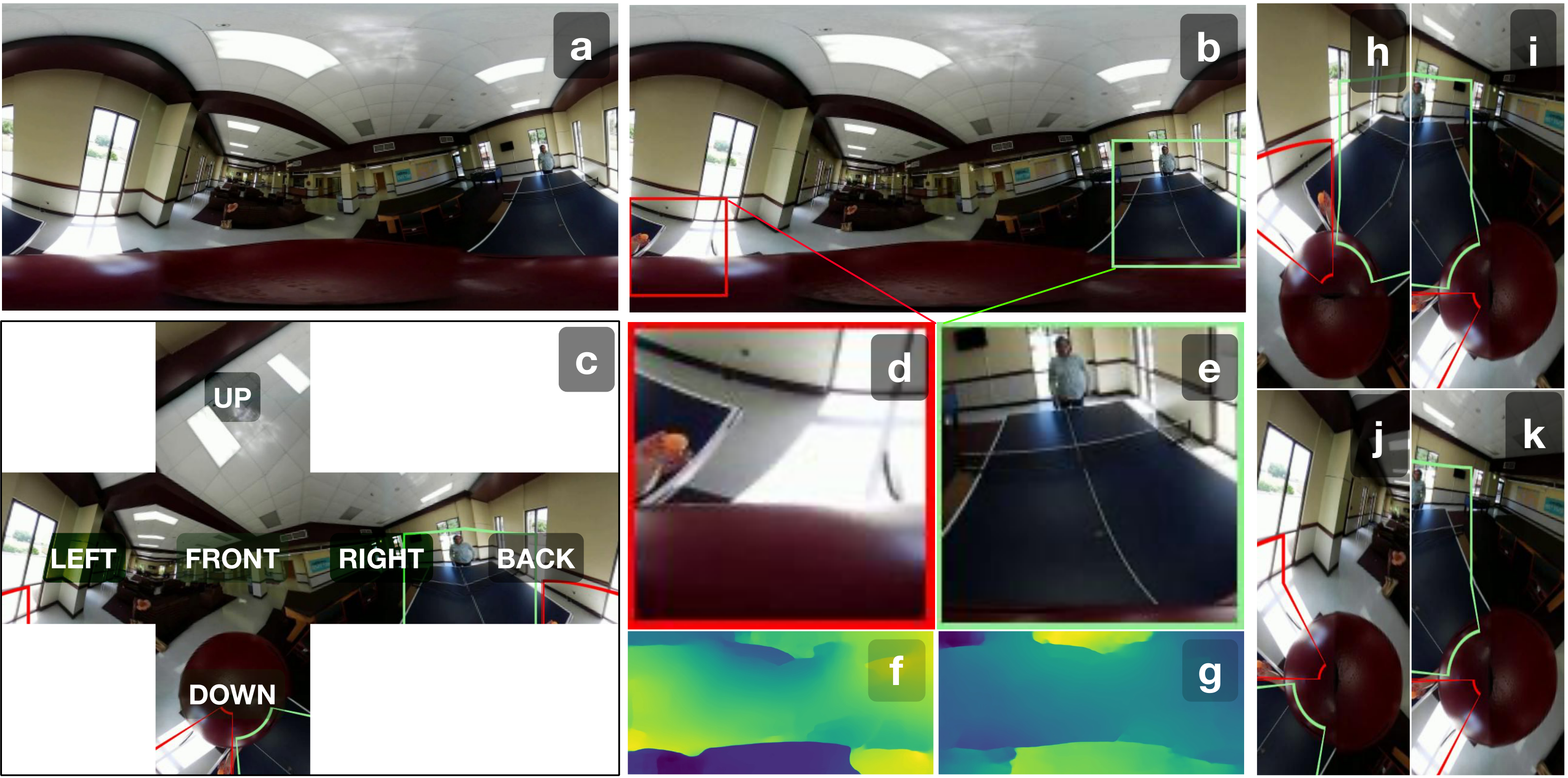}
\end{center}
\vspace{-0.1in}
   \caption{Sample video frames from EgoK360 Dataset. (a,b) Consecutive frames ($ I_i,I_{i+1}$) for action ``Serve" from ``Ping-Pong" activity in equi-rectangular projection. (c) Cubemap projection of (b) showing six different cubic faces. (d,e) Cropped section of wearer showing action `serving' (red box in (b)) and front-view from wearer's perspective (green box in (b)). (f,g) Optical flow ($\vec u_i,\vec v_i$). (h,i,j,k) Normal field-of-view for front-down, back-down, left-down and right-down.}
\label{fig:samplesfigure}
\end{figure}


In this paper, we propose a novel 360 egocentric human activity recognition (EgoK360) dataset. The EgoK360 dataset is inspired by action recognition datasets such as UCF-101~\cite{ucf101}, HMDB-51~\cite{hmdb} and Kinetics~\cite{i3d}. Our EgoK360 dataset contains three different types of actions: Person-Person, Person-Object, and Singular actions. These categories of actions can be described in the following manner. Person-Person actions involve two or more people  interacting with each other such as hugging and speaking with someone; Person-Object actions refer to a person interacting with some objects such as picking up something or moving something from one location to another; Singular person actions involve a single person performing some actions independent of others such as reaching towards something or combing hair.
In this paper we perform experiments with two popular action classification deep nerual networks on our introduced EgoK360 dataset, \textit{i.e.}, two-stream network \cite{twostream} and Inflated 3-Dimensional network (I3D) \cite{i3d}. In the following sections, we present the related work, datasets, experimental results and conclusions. 

\section{Related Work}
\label{sec:format}
Action recognition datasets such as HMDB \cite{hmdb}, UCF101 \cite{ucf101} and Kinetcs \cite{kinetics} are widely used in literature. They are captured by perspective cameras (single field-of-view) and have limitations in terms of applications. Singh \etal~\cite{egocentricvideos} use a novel dominant motion feature derived from optical flow for egocentric action recognition and also propose a convolutional neural networks (CNN) \cite{firstperson} for end-to-end training. Xia \etal~\cite{robots} present a framework to analyze RGBD videos captured from a robot for activity recognition. Lee \etal~\cite{utego} present an egocentric video summarization approach by identifying important people and object in the video. Two-stream network is the popular architecture in literature for action recognition, such as Two-stream Convnet \cite{twostream} and Inflated 3D ConvNet (I3D) \cite{i3d}. I3D architecture is the state-of-the-art in the two-stream genre for action recognition.



In recent years, a few 360$^{\circ}$ datasets \cite{\comment{vr4,}vr3, vr2, sp2, vr5, cubepadding,\comment{vr1,vr4,}sd2} appeared in the applications such as autonomous driving, human-computer interaction, virtual reality, and others. However, they are target to different applications other than  Egocentric Activity Recognition in 360$^\circ$ field-of-view (EAR360). \comment{Do we need to explain how different they are? YES}

Meanwhile, in the egocentric action recognition field, popular datasets such as \comment{Epic Kitchens \cite{kitchen},} EgoHands \cite{egohands}, EGTEA Gaze+ \cite{beholder} are perspective video datasets with a person interacting with an object or another person. Similarly, large-scale datasets such as Charades-Ego \cite{charades} contains both the first-person and third-person videos.\done{[change to some people present add egocentric as first person aka]} Pirsiavash \etal~\cite{dailyliving\comment{, task, act, disc, story}} present an egocentric dataset for understanding activities and the context in the video. However, all these datasets in literature are only limited to perspective videos.


\vspace{-0.05in}

\section{EgoK-360 Dataset}
\label{sec:pagestyle}
Our EgoK-360 dataset contains activity classes that represent all three categories, \textit{i.e.,} Person-Person, Person-Object, and Singular actions. There are a few differences in the video content compared with other datasets because of the properties of Egocentric 360 videos. Given that the footage encompassing the dataset captured from an egocentric perspective, the Person-Person actions would involve the interaction between the wearer and other people. This differs with Person-Person actions captured in the traditional third-person perspective cameras. Likewise, a Person-Object action such as bouncing a ping-pong ball on a table would only be identified when the particular action was performed by the wearer. Most action classes in EgoK360 are in the Person (singular) category because the egocentric perspective inherently privatizes the action or content recorded. 

\done{REWRITE THIS PARAGRAPH!}
The 360 field-of-view naturally makes everything egocentric in EgoK360. Egocentric actions entirely depend on the field-of-view where a wearer (first-person) is engaged. Meanwhile, the rest of fields-of-view that he/she is not engaged are irrelevant for action recognition.  
The significant contribution to action recognition is the wearer’s egocentric view and his/her engagement in actions. 

\begin{figure}[htb]
\begin{center}
\includegraphics[width= \linewidth]{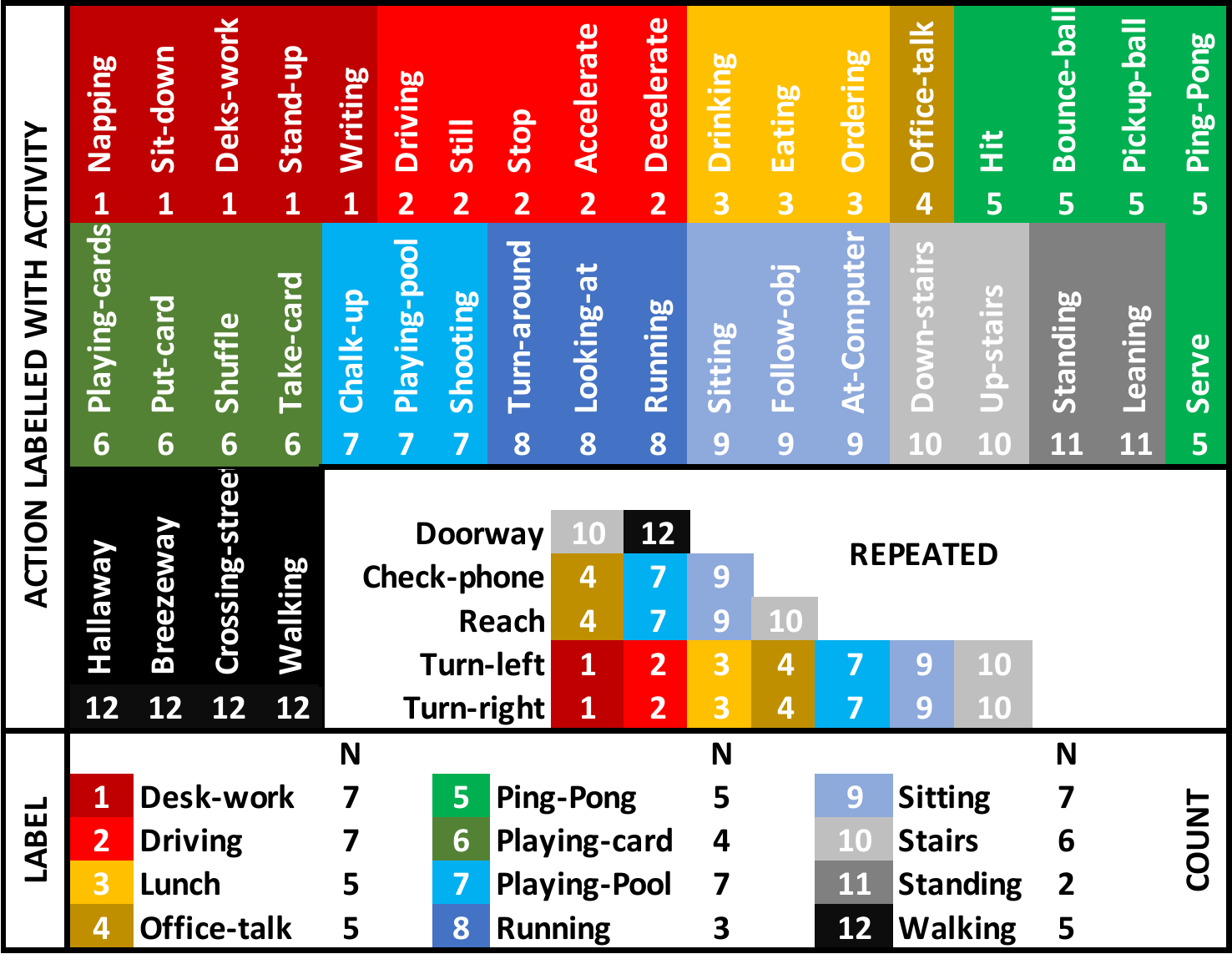}
\end{center}
\vspace{-0.1in}
   \caption{All activities and action classes in EgoK-360. Actions are colored and numbered with corresponding activity. The same actions may appear in different activities.}
\label{fig:instances}
\end{figure}

\subsection{Activity/Action Classes}
We show action/activity instances of EgoK360 in Fig.~\ref{fig:instances}. Our dataset contains 12 activities and 45 actions, collectively making 63 activity-action unique cases. An activity is defined as collections of shorter actions. For example, an activity `driving' is composed of actions such as accelerate, decelerate, idle, stop, driving, turn-left and turn-right. Action classes such as turn-left, turn-right, reach, doorway and check-phone are frequently occurring actions. However, there is a significant difference in these actions depending upon the category of activity. For example, turn-left in driving is completely different than turn-left in activity office-talk. We collected 127 videos with approximately 11 minutes each. 

\subsection{EgoK-360 Characteristics}
The EgoK-360 dataset has its uniqueness of 360 fields-of-view, egocentric and kinetic properties. We discuss the following characteristics of EgoK-360 dataset in terms of its diversity, statistics and properties.

\noindent \textbf{Diversity.} Our EgoK-360 dataset contains common different activities in daily life. Around 11\% of actions (such as turn-left, turn-right, reach, check-phone and doorway are frequent actions) are overlapping actions. Activity such as desk-work, driving, playing-pool and running have the most number of actions. Activity such as standing has the least number of actions. 

\noindent \textbf{Properties.} We present sample frame in Fig.~\ref{fig:samplesfigure}. The dataset is a collection of videos from a 360 camera projected on the 2D plane using equirectangular projection, as shown in Fig.~\ref{fig:samplesfigure} (a and b). The frames size is 640x320. Frames exhibit huge distortion as shown in Fig.~\ref{fig:samplesfigure} (b-c, h-k) using red and green bounding box), making it challenging for regular convolution. We calculate optical flows using FlowNet~\cite{flownet2}.

\section{Experiment}

\label{sec:typestyle}
We conduct our experiments using two-stream and I3D networks. We implement two-stream architecture with  resnet-101 model pre-trained on UCF101 which outperforms state-of-the-art I3D model. EgoK360 exhibits  complexity of spherical representation of 360$^{\circ}$ video on 2D plane (equirectangular projection) which makes challenge for these models to prioritize a significant field-of-view responsible for the wearer's engagement in certain actions and makes difficult to train. Therefore, 3D-representation of the video does not perform well in the 360 environment.

\begin{figure}[htb]
\begin{center}
\includegraphics[width= \linewidth]{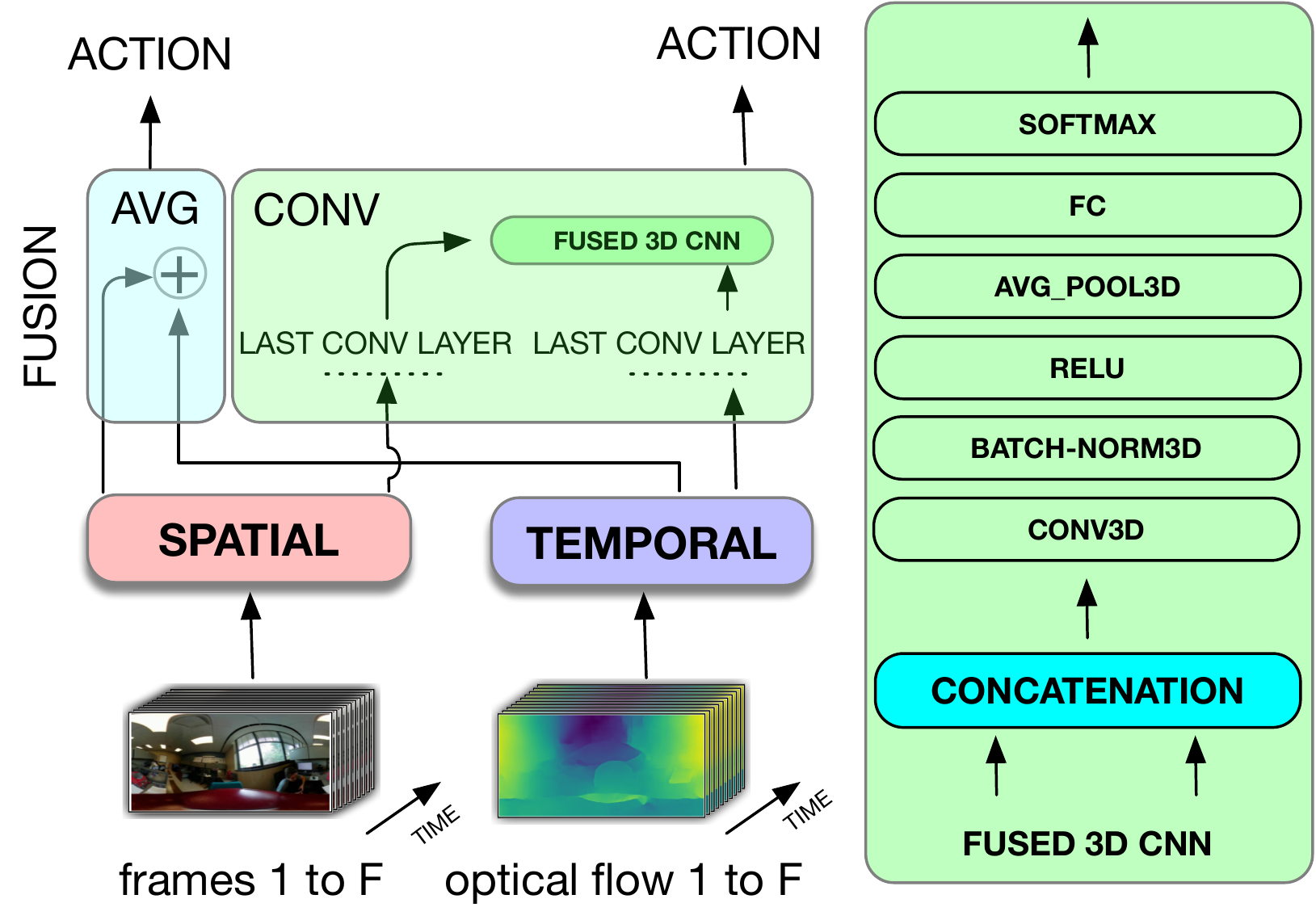}
\end{center}
\vspace{-0.1in}
   \caption{Spatio-temporal architecture for action/activity classification. We implement resnet-101 and I3D architecture. Average and convolution fusion are adopted. For average fusion we simply average probabilities of two networks and map them into single probability. For convolution fusion, we concatenate (depth-wise) output from last convolution layer and feed to the convolution module.}
\label{fig:architecture}
\vspace{-0.2in}
\end{figure}

\subsection{Implementation Details}
We adopt the network architecture as shown in Fig.~\ref{fig:architecture} in our experiments. Our model inputs are consecutive frames. Videos are down-sampled in the rate of 10 fps. We calculate optical flows beforehand using FlowNet~\cite{flownet2}. We adopt the two-stream and I3D architectures with average and convolution fusion. For two-stream architecture, video is represented as 2D inputs with $[N \times F_c \times H \times W]$ dimensions. For I3D architecture, video is represented as 3D input with $[N \times C \times F \times H \times W]$ dimensions, where $F_c = F \times C$. Here $F$ is the number of frames, $N$ is the batch-size, $C$, $H$ and $W$ are channel, height and width of the frames.

\subsection{Two-stream Architecture}
Residual learning framework \cite{resnet} provides convenient optimization and rapid high accuracy as network becomes deeper. With this in mind, we change the two-stream architecture by replacing the spatial and temporal network with resnet-101 model pre-trained on UCF101. We use size of 10 to stack frames in sequence for both spatial and temporal networks. This brings the channel size changing from 3 to 20 in the temporal network and to 30 in the spatial network. The resnet-101 requires input as 3 channel images. To fix this, we use the method in the cross-modal learning \cite{crossmodal}. We do not observe better results  compared to UCF101 implemented with the same architecture, which achieves at least 80\% accuracy.

\subsection{I3D Architecture}
We implement I3D architecture as proposed in \cite{i3d}. I3D architecture relies on 3D receptive fields for video representation. Spatial and temporal network receive an input of 3 and 2 for the channel size respectively, along with depth of 10. The original idea in \cite{i3d} using the entire video as one training sample. However, we do not achieve  performance increase on EgoK-360 dataset. We run  experiments with the depth of 10 which is the optimal for our case.

\begin{table*}

  \centering
\begin{adjustbox}{width=0.8\linewidth}

    \begin{tabular}{|l|r|r|r|r|r|r|r|r|}
    \hline
    
    \multicolumn{1}{|c|}{} & \multicolumn{8}{|c|}{Network Mode}\\
    
    
        \cline{2-9}

          \multicolumn{1}{|c|}{Architecture} & \multicolumn{2}{c|}{flow} & \multicolumn{2}{c|}{rgb} & 
          
          \multicolumn{2}{c|}{avg\_fused} &
          
          \multicolumn{2}{c|}{conv\_fused}\\
          \cline{2-9}
          \multicolumn{1}{|c|}{} & \multicolumn{1}{l|}{Activity} & \multicolumn{1}{l|}{Action} 
          & \multicolumn{1}{l|}{Activity} & \multicolumn{1}{l|}{Action} 
          & \multicolumn{1}{l|}{Activity} & \multicolumn{1}{l|}{Action} 
          & \multicolumn{1}{l|}{Activity} & \multicolumn{1}{l|}{Action}  \\
          
    \hline
    \multirow{1}{*}{Resnet}
    & 73.94 & 61.09 & \textbf{77.05} & 58.22 & 76.53 & 62.44 & 74.71 & 56.87\\
    \hline
    \multirow{1}{*}{I3D}
    & 57.24 & 43.4  & 74.13 & 55.31 & 68.74 & 50.88 & \textbf{74.47} & 56.63 \\
    
    \hline 
    \end{tabular}%
    
    \end{adjustbox}
   \bigskip
   \vspace{-0.2in}
  \caption{Experimental results of EgoK360 datasets on two-Stream (modified version with trained Resnet-101 on UCF101) and I3D Architecture. (\textit{Top accuracy in bold})}
  \label{tab:addlabel}
 
\end{table*}%

\subsection{Fusion}
In this paper we use both average and convolution fusion techniques. In average fusion, we take an average of probabilities from the last layers. For convolution fusion, we implement convolution-module inspired by \cite{resnet}. We use the output from the last convolution layer and concatenate the features which later fed into the fusion convolution module. We freeze our spatio-temporal module and train the fusion layer. We can also train the spatio-temporal network along with the fusion layer.

\subsection{Results}
\begin{figure}[htb]
\begin{center}
\includegraphics[width=9.0cm,height=5.0cm]{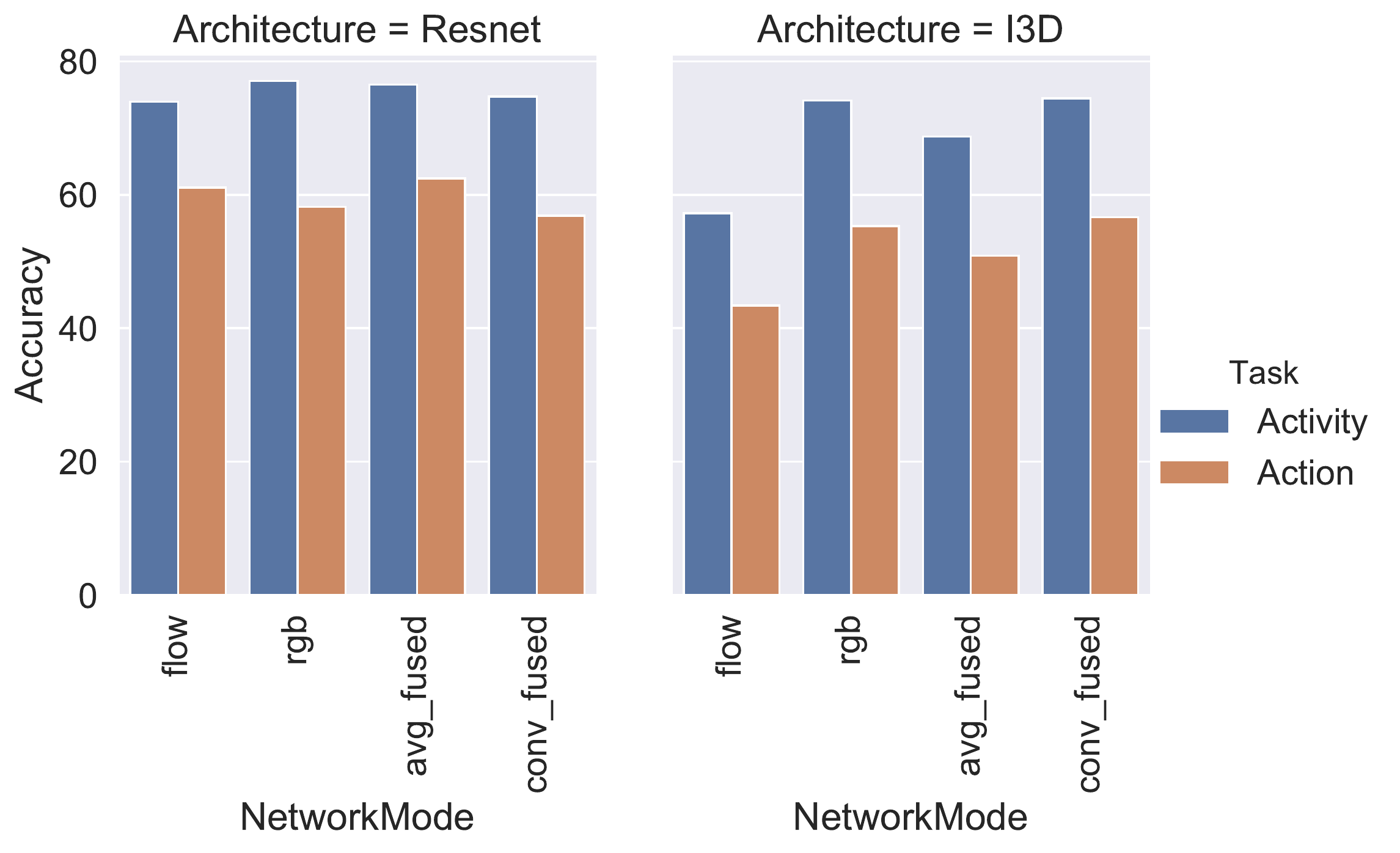}
\end{center}
    \vspace{-0.1in}
   \caption{Visualization results of Table \ref{tab:addlabel}. 
   The figure shows accuracy for each architecture and classification mode (activity \textit{vs} action). Overall 'action' classification is better in resnet-101. We observe convolution fusion in resnet architecture and I3D makes significance difference in flow (temporal) and rgb (spatial) stream. Similarly, temporal-stream performs better in resnet compared with I3D architecture. Conv\_fusion has same effect on both cases where as avg\_fusion comparatively improve resnet architecture. In general resnet architecture shows consistent metrics relative to I3D arhcitecture.}
\label{fig:visualization}
\end{figure}

We present our experimental results in Table \ref{tab:addlabel} and visualization in Fig.~\ref{fig:visualization}. We observe that activity classification accuracy is higher than action classification accuracy as shown in Table-\ref{tab:addlabel}. The range of activity and action classification accuracy in I3D architecture is  higher than in two-stream architecture.

The average fusion is remarkably better than the convolution fusion in our case. This quantitative results can be explained using Fig-\ref{fig:visualization} visualization. We observe interesting results on two different architectures. Fusion techniques make a huge difference in action/activity classification in resnet whereas spatial and temporal streams have significant differences in I3D architecture. From Fig.~\ref{fig:visualization} we can infer that convolution fusion performs better whenever two streams have significant gap in accuracy.

\begin{figure}[htb]
\begin{center}
\includegraphics[width=8.5cm,height=6.2cm]{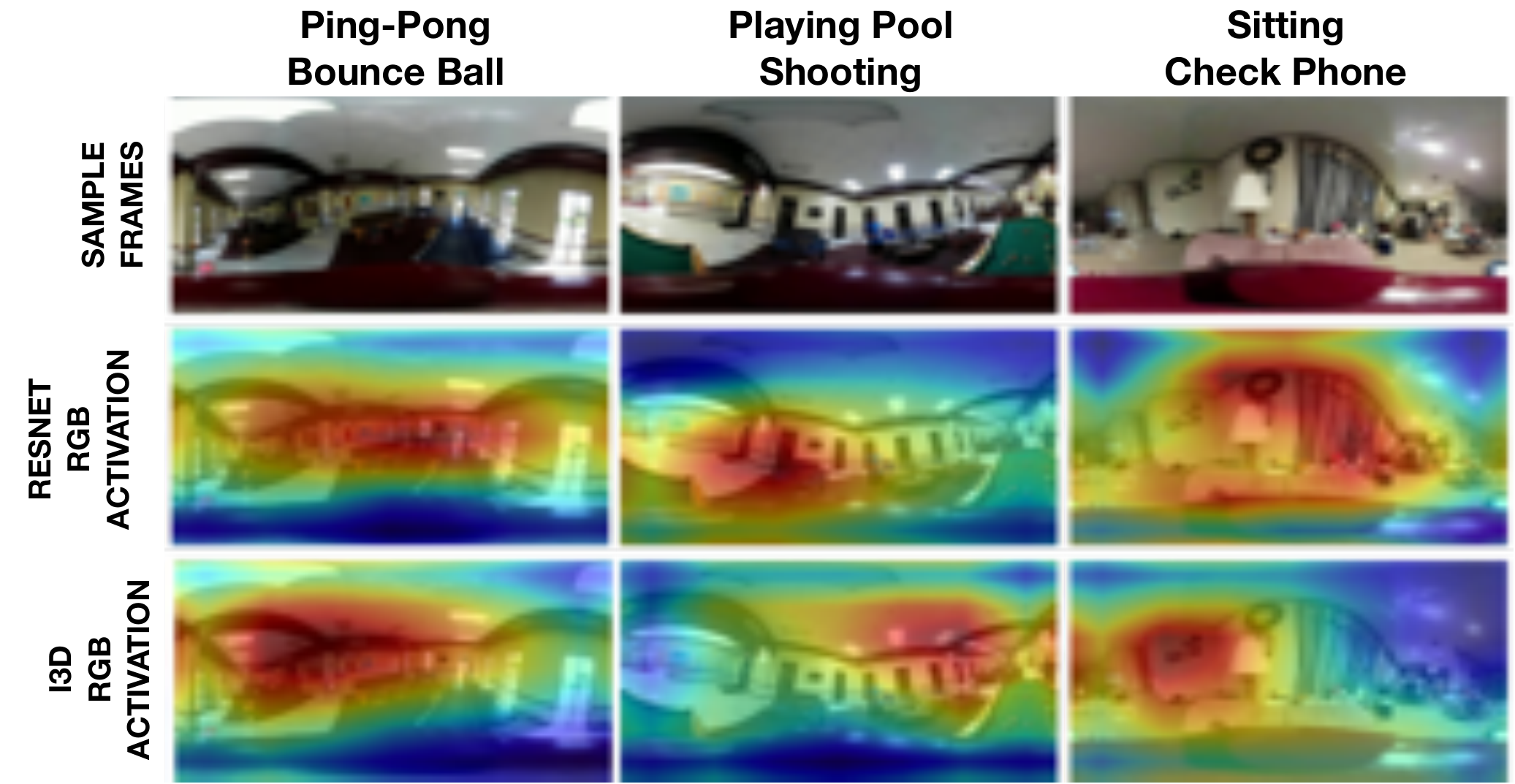}
\end{center}
\vspace{-0.1in}
   \caption{Activation map showing salient features learned by the spatial network. The top row shows RGB frames, the second row represents activation map from two-stream networks, and the bottom row shows the activation map from I3D model.}
\label{fig:activations}
\end{figure}
We also investigate how well this two-stream architecture generalizes with our dataset. We use the technique presented in \cite{cam} to visualize the activation map. We show the activation map with a randomly selected action in Fig.~\ref{fig:activations}. We derive these activation maps from the I3D and two-stream spatial network.

These activation maps represent salient features learned by the model. The model infers most edges as trivial regions, as the dataset has massive distortion near edges. We can visually inspect and analyze this behavior. For example, in the Fig.~\ref{fig:activations} activity Bounce\_ball (playing Ping-Pong activity), the salient features are away from the actual region where a person wearing a camera is bouncing a ball. This region lies on the left-bottom-corner and has massive distortion. It is nearly impossible to judge the meaning of these activation maps accurately. However, if we carefully inspect the salient features learned by both architectures, we can conclude that the model is inferring the action classification task from other features rather than salient features as expected. The reason behind this poor response of the model is due to the na\"ive convolution, which is not rotation invariant. Features on EgoK360 have different spatial properties depending upon the position in equirectangular plane. This can be improved with techniques such as~\cite{ktn, sphconv}. 

\section{Conclusion}
This paper introduces EgoK360 dataset with annotations of 63 unique activity and action classes. This dataset is challenging because of distortion, wide field-of-view and activities/actions properties. We implement two popular two-stream architectures in the experiments. We modify the two-stream convents architecture by replacing each stream with resnet-101. It outperforms state-of-the-art I3D architecture. EgoK-360 is the first to address egocentric activity recognition in 360 environment. We believe EgoK360 dataset will be beneficial to the EAR360 research.

\vspace{0.2in}

\noindent \textbf{Acknowledgements:} This research was partially supported by NSF CSR-1908658, NeTS-1909185, and gift donation from Cisco Inc. This article solely reflects the opinions and conclusions of its authors and not the funding agents.


\bibliographystyle{IEEEbib}
\bibliography{strings,refs}

\end{document}